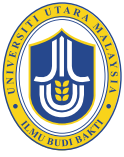

# PARAMETRIC FLATTEN-T SWISH: AN ADAPTIVE NON-LINEAR ACTIVATION FUNCTION FOR DEEP LEARNING

**[1]Hock Hung Chieng, [1]Noorhaniza Wahid & [2]Pauline Ong**
[1]Faculty of Information Technology and Computer Science,
Universiti Tun Hussein Onn Malaysia, Malaysia

[2]Faculty of Mechanical and Manufacturing Engineering,
Universiti Tun Hussein Onn Malaysia, Malaysia

*Corresponding author: hi160029@siswa.uthm.edu.my
nhaniza, ongp@uthm.edu.my*

**ABSTRACT**

Activation function is a key component in deep learning that performs non-linear mappings between the inputs and outputs. Rectified Linear Unit (ReLU) has been the most popular activation function across the deep learning community. However, ReLU contains several shortcomings that can result in inefficient training of the deep neural networks, these are: 1) the negative cancellation property of ReLU tends to treat negative inputs as unimportant information for the learning, resulting in performance degradation; 2) the inherent predefined nature of ReLU is unlikely to promote additional flexibility, expressivity, and robustness to the networks; 3) the mean activation of ReLU is highly positive and leads to bias shift effect in network layers; and 4) the multilinear structure of ReLU restricts the non-linear approximation power of the networks. To tackle these shortcomings, this paper introduced Parametric Flatten-T Swish (PFTS) as an alternative to ReLU. By taking

ReLU as a baseline method, the experiments showed that PFTS improved classification accuracy on SVHN dataset by 0.31%, 0.98%, 2.16%, 17.72%, 1.35%, 0.97%, 39.99%, and 71.83% on DNN-3A, DNN-3B, DNN-4, DNN-5A, DNN-5B, DNN-5C, DNN-6, and DNN-7, respectively. Besides, PFTS also achieved the highest mean rank among the comparison methods. The proposed PFTS manifested higher non-linear approximation power during training and thereby improved the predictive performance of the networks.

## INTRODUCTION

In recent years, deep learning has brought tremendous breakthroughs in artificial intelligence (AI). Such astonishing advancements are due to these factors: the availability of the massive amount of data, powerful computational hardware such as Graphic Processing Units (GPUs), and deep learning models. This study focuses on the latter breakthrough.

A deep learning model is a composition of several functions that mimic how the brain works (Hassabis et al., 2017; Chen et al., 2019). One of the important elements that contribute towards its learning power lies in the non-linear activation function (Ciuparu et al., 2019). Since the revival of deep learning in 2012, Rectified Linear Unit (ReLU) (Nair & Hinton, 2010) has been the most popular and commonly used non-linear activation function across diverse deep learning models. ReLU holds two major advantages. First, ReLU function is simple and easy to implement in any deep learning model (Lin & Shen, 2018). It keeps the positive inputs and discards the negative inputs. This structure of ReLU substantially speeds up the learning and reduces the computational cost. Second, the non-saturation property of ReLU in the positive region ensures smooth gradient flow and avoids vanishing or exploding gradient problems (Nair & Hinton, 2010). Unlike classical methods such as Sigmoid and Tanh, the saturation properties both at the negative and positive regions further impede the gradient flow during the model training.

Despite its superiority, ReLU also comes with several shortcomings. The negative region of ReLU treats negative input as useless representation (or information) for the learning (Qiu et al., 2018). In fact, several previous studies have found that making use of negative inputs can substantially improve the predictive performance (Qiu et al., 2018; Chieng et al., 2018; Maas et al., 2013). In addition, ReLU is a predefined or non-adaptive activation function and therefore is unable to learn and adjust according to the received inputs. Some studies considered the use of adaptive strategy in the activation functions to improve the network's robustness, flexibility, adaptivity, generalisation power, and predictive performance (Qiu et al., 2018; Qian et al., 2018; He et al., 2015). Another limitation of ReLU comes from its inherent function structure. ReLU is a multilinear system that is formed by two linear lines (Laurent & Von Brecht, 2018). To further explain this point, the ReLU network is simply a model that performs a linear combination of input values (Yarotsky, 2018). Consequently, ReLU networks have poorer non-linear approximation power. Figure 1 explains the non-linear approximation power of the ReLU and Swish networks pictorially. Figure 1(a) reveals that the network with ReLU struggles to approximate the polynomial function, while the network with Swish in Figure 1(b) can fit the polynomial function better. This implies that the activation function that possesses a certain degree of curvature may endow a higher non-linear capability for the networks to learn complex non-linear representation. In addition, the mean activation of ReLU is not zero-centred. Since ReLU only produces non-negative values as output, the output mean is always larger than zero. This leads to a problem commonly known as bias shift effect (Clevert et al., 2016), which makes the training challenging.

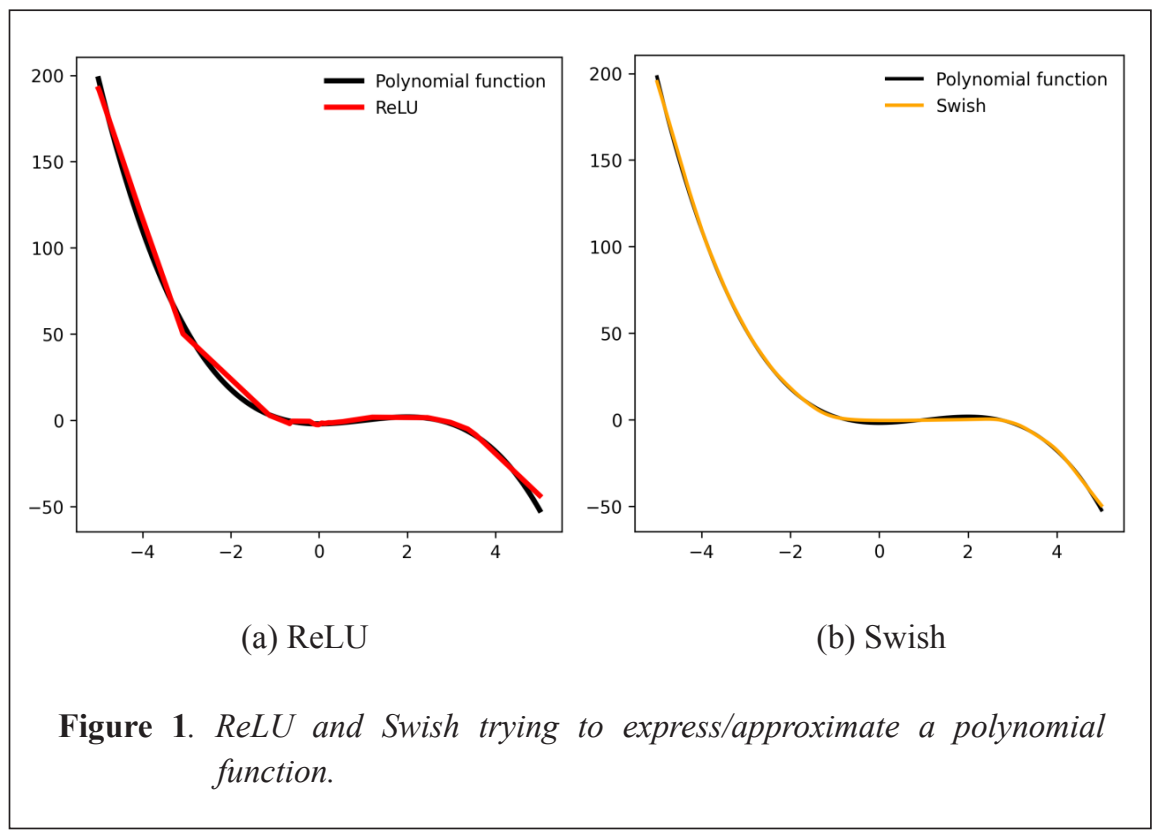

(a) ReLU          (b) Swish

**Figure 1**. *ReLU and Swish trying to express/approximate a polynomial function.*

This study aims to tackle the shortcomings of ReLU by introducing an adaptive non-linear activation function called Parametric Flatten-T Swish (PFTS). This paper also seeks to extend the previous idea, which has not been covered in the work of Chieng et al. (2018). This work provides a much deeper understanding and insight into the performance of the proposed activation function from a different perspective. The proposed activation function will be compared empirically with nine existing activation functions on image classification tasks. The rest of the paper is structured as follows. Related Works discuss the previously existing methods. The Proposed Method section presents the proposed activation function. Then, the Experimental Setup section will explain the procedure of evaluating the performance of the proposed activation function. The analysis of the results will be discussed in this section. Finally, the conclusion and potential future work are drawn at the end of this paper.

## RELATED WORKS

Non-linear activation function is one of the key components that substantially influence the behaviour of deep learning models. In the past decade, numerous non-linear activation functions have been introduced trying to facilitate the performance of deep learning models in diverse real-world applications. Generally, non-linear activation functions can be classified into two basic approaches: predefined activation functions and parametric activation functions.

**Predefined Activation Functions**

Predefined activation functions are also known as fixed activation functions. These activation functions do not change their innate nature and characteristic along with the network training. Their functions' shape remains unchanged before and after the training. The most commonly used fixed activation function is ReLU. ReLU is defined by Equation 1:

$$ReLU(x) = \begin{cases} x, & if\ x \geq 0 \\ 0, & if\ x < 0 \end{cases} \quad (1)$$

where $x$ is the input from the previous layer. ReLU discards negative signals and keeps positive signals unchanged. Besides, its derivative is constant 0 at the negative region and 1 at the positive region. The simplicity of ReLU makes the training of deep learning models feasible and less computationally expensive. On the other hand, the negative cancellation property of ReLU could potentially stall the training due to the fact that a vast number of neurons in the network have become inactivated permanently. To overcome this, Leaky ReLU (LReLU) (Maas et al., 2013) was introduced. LReLU allows a small slope at the negative region to activate the negative values, while the positive part remains similar to ReLU.

Another work proposed Exponential Linear Unit (ELU) (Clevert et al., 2016) to deal with the bias shift effect of ReLU. Since ReLU removes the negative inputs, it means activation is always positive. Thus, it will slow down the training. ELU uses exponential property at the negative region to create a smooth curvature to push the mean activation towards zero. This unique property further enables the activation function to behave like batch normalisation (BN) (Ioffe & Szegedy, 2015), which results in better generalisation power and accelerate learning. ELU was later improved to Scaled ELU (SELU) (Klambauer et al., 2017). As its name would suggest, SELU is the scaled variant of ELU. It was originally used in Self Normalise Neural Network (SNN) to keep the activation function normalised throughout the training. This unique property of SELU tackles the bias shift effect and high variance in the network layers simultaneously. As a result, SELU leads to faster convergence and improves model predictive performance as compared to ReLU and ELU (Liu et al., 2019).

All the above reviewed non-linear activation functions are based on the piecewise system, in which more than one function are composited in an activation function. The work of Ramachandran et al. (2018) proposed a non-piecewise activation function based on the Sigmoid function called Swish. It is simply a scaled Sigmoid function. The function curve is smooth, unbounded at the positive region, and bounded at the negative region. The approach proves the feasibility of smooth activation function in training the deep learning model after a long abandonment of sigmoidal activation functions. Swish is superior in terms of expressive power and fast convergence. It has been applied in diverse deep learning applications and proved to be more effective than ReLU and other activation functions (Tripathi et al., 2019; Jinsakul et al., 2019; Wang et al., 2019).

In all these methods of Maas et al. (2013), Clevert et al. (2016), Klambauer et al. (2017), and Ramachandran et al. (2018), the expressive power and the flexibility are still limited even though the scholars could make the training of deep learning models possible. In fact, several recent studies preferred a more dynamic approach, where the parameterisation approach is used to endow greater flexibility and expressivity for the deep learning models.

**Parametric Activation Functions**

Parametric, in this case, is sometimes also termed as trainable or adaptive. This category of activation function has gained much attention in recent years after it has been found to be closely related to network performance. Parametric activation functions fine-tune a certain parameter automatically along with the network's weights and biases. Unlike the predefined one, the shape of the parametric activation functions is learned and changes consistently during training.

Adaptive piecewise linear (APL) (Agostinelli et al., 2015) can be considered as the first parametric activation function introduced to deep learning. As the name suggests, APL is a multilinear system. APL is adaptive in the way that it learns the slopes of its local linear segments to determine the optimal location of the hinge points. The use of APL in deep learning models leads to greater dynamicity and expressive power. A past experiment demonstrated the distinguished performance of APL over the non-parametric activation functions on image classification tasks (Agostinelli et al., 2015).

Instead of learning the multiple slopes, Parametric ReLU (PReLU) (He et al., 2015) improved the LReLU by learning the slope at the negative region automatically. The advent of PReLU has led to a significant breakthrough in visual recognition task and became the first deep learning approach that surpasses the human-level performance. This has further aroused the attention of the community on the impact of the parametric activation function towards the network's performance. Upon this similar concept, the parametric idea was then also applied to ELU and named as Parametric ELU (PELU) (Trottier et al., 2017). Instead of learning the negative region alone, PELU adds a learnable parameter at its positive part so the positive slope can also be learned. The work revealed that inserting additional parametric components might generally improve network flexibility and predictive performance.

While most of the parametric approaches are focusing on learning either the negative or positive, or both regions, Flexible ReLU (FReLU) (Qiu et al., 2018) uses a slightly different parametric approach. FReLU learns the activation function as a whole without having to manipulate its original shape or any slope in the function. FReLU allows the function to be adaptive vertically by learning the hinge point. This idea was proposed primarily to deal with the inherent shortcomings of ReLU, which are negative cancellation and bias shift effect. FReLU has demonstrated its distinguished performance over other methods on image classification tasks. This shows credible evidence that negative activation and adaptivity are of important elements for activation function to enhance deep learning models.

**Criteria of Good Activation Function**

The above non-linear activation functions only focus on solving part of the inherent problems of ReLU. However, a general criterion for a good activation function can also be sorted out from the previously proposed approaches. Here, four possible criteria are summarised.

*Negative activation*

As stated previously, almost all the activation functions introduced after ReLU own negative activity property. In general, this class of activation functions has better predictive performance than ReLU.

*Zero-centred mean activation*

Activation functions such as ELU, SELU, and FReLU bring mean activation closer to zero. This property is highly important for deep learning models to mitigate the bias shift effect in the network layers through normalising the input values. As a result, the learning speed is accelerated.

*Parameterisation*

Parameterisation is also known as adaptivity or trainability. Unlike the fixed ones, the parametric activation functions in network layers learn their activation independently together with the weights and biases of the network during the training. This provides greater flexibility as well as enhances the non-linear representation ability of the network.

*Locally non-linear*

Activation functions like ReLU, LReLU, APL, and PReLU are multilinear in structure. Meanwhile, ELU, SELU, PELU, and Swish contain locally non-linear regions. The previous works have demonstrated that the locally non-linear activation functions consistently outperformed multilinear activation functions across the different network architectures on visual recognition tasks. This is proven by a recent theoretical study of Ohn and Kim (2019) that the locally non-linear region can promote better expressivity and non-linear approximation capability.

## PROPOSED METHOD

The previous work of Chieng et al. (2018) proposed a Flatten-T Swish (FTS) activation function to deal with the negative cancellation property in ReLU. The experimental results revealed that FTS outperformed ReLU consistently in all five network setups on image classification tasks. Mathematically, FTS is defined in Equation 2:

$$FTS(x) = \begin{cases} \dfrac{x}{1+e^{-x}} + t, & if\ x > 0 \\ t, & if\ x < 0 \end{cases} \qquad (2)$$

where *x* is the input from the previous layer and *t* is a predefined hyperparameter. *t* is suggested to be fixed at -0.20 according to the original work. Figure 2 shows the plot of FTS. Intuitively, FTS can be viewed as a scaled Sigmoid function for the region *x*. It is worth noting that the symbol *e* in Equation 2 is an expression for the exponential function and it is also used consistently throughout the relevant activation functions in this paper. ≥ 0, while the parameter *t* provides a constant negative output for the region *x* < 0.

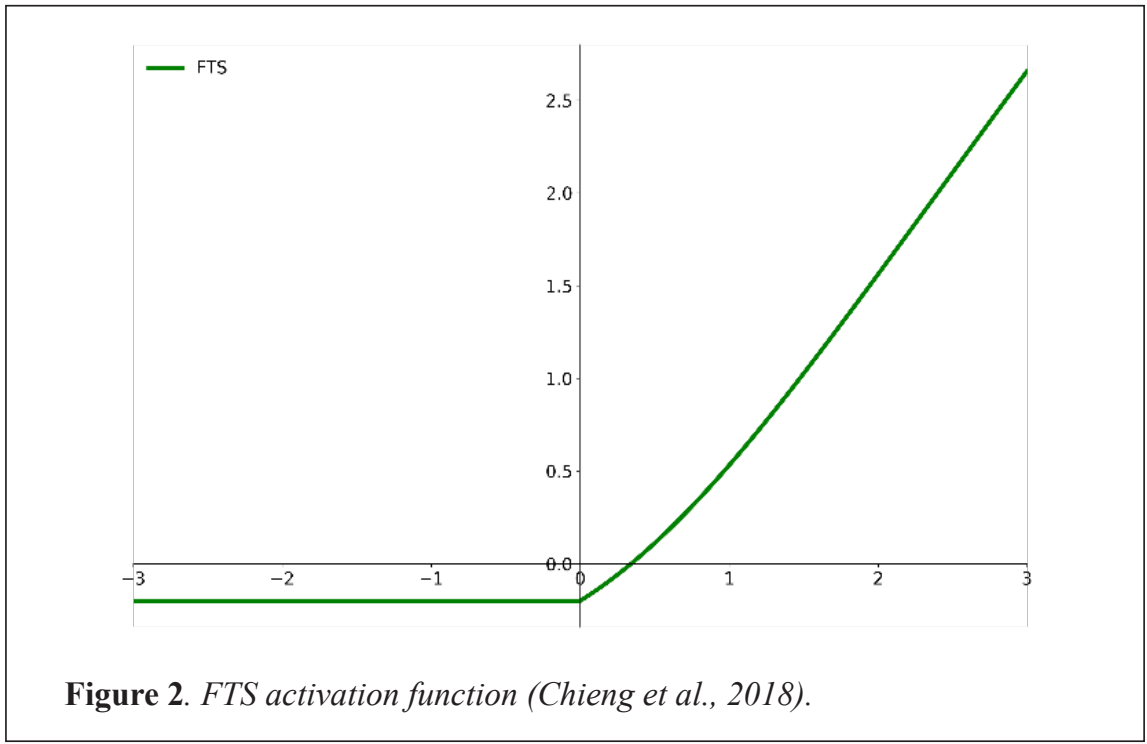

**Figure 2**. *FTS activation function (Chieng et al., 2018).*

FTS is a predefined activation function, and therefore, it does not provide additional advantages such as dynamicity, flexibility, and non-linear representation ability for the networks. For this reason, a Parametric Flatten-T Swish (PFTS) is proposed to overcome these limitations. Formally, PFTS is defined by Equation 3:

$$PFTS(x) = \begin{cases} \dfrac{x}{1+e^{-x}} + t_{train}, & if\ x \geq 0 \\ t_{train}, & if\ x < 0 \end{cases} \qquad (3)$$

where *x* indicates the input to the activation function and $t_{train}$ denotes the adaptive parameter that allows it to be learned from the training. The initialisation value for $t_{train}$ $t_{train}$ is set to be -0.20 before the training. While most of the existing parametric approaches are focusing on learning the slope(s) of the function segments, PFTS learns the whole activation function as in FReLU and thus the function shape is maintained. Specifically, PFTS learns the hinge point $t_{train}$ of the function. This dynamic component allows PFTS to vertically fluctuate freely to discover the best value for $t_{train}$. Thus, different network layers will give different activation response. Figure 3 shows the adaptivity direction of PFTS. In terms of the function structure, PFTS shares the same function shape as FTS. More specifically, the right region of the functions is essentially similar to Swish, and the left region maintains as ReLU-like property.

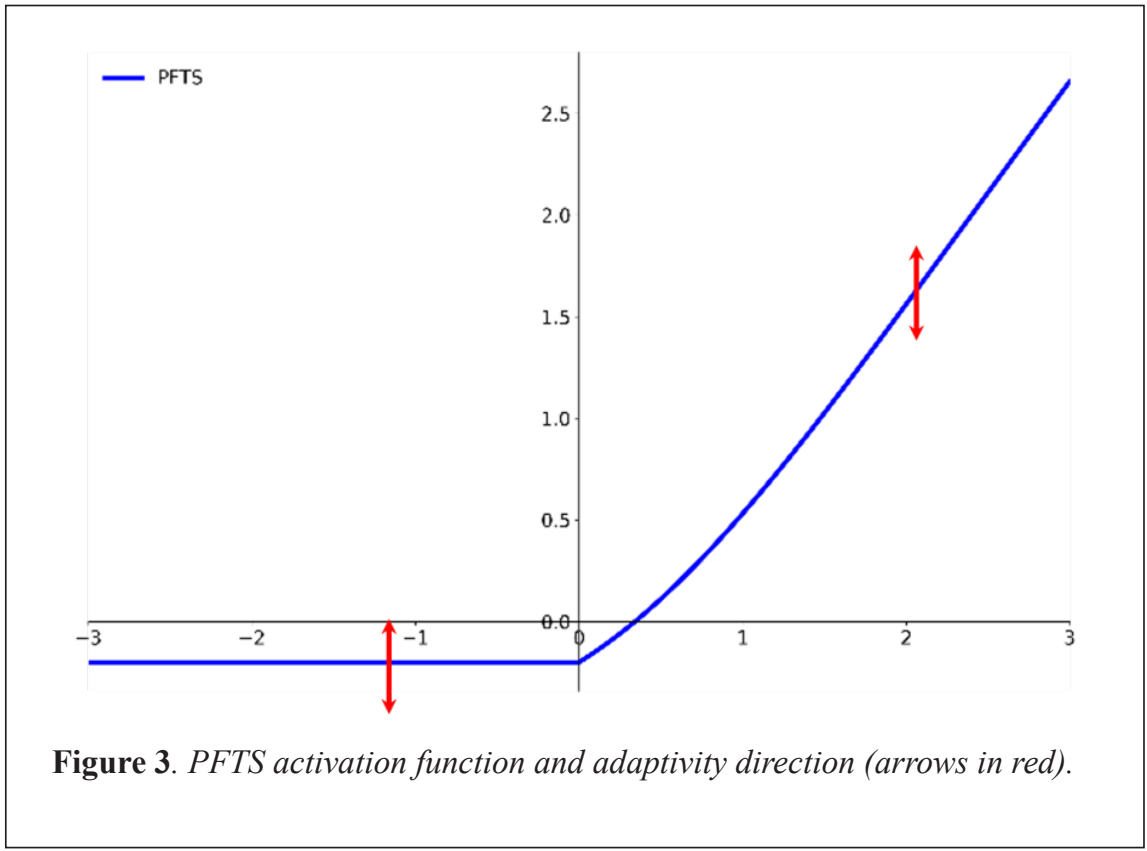

**Figure 3**. *PFTS activation function and adaptivity direction (arrows in red).*

Deep Neural Network (DNN) updates its parameters such as weights and biases through the backpropagation optimisation technique. The optimisation technique simply computes the derivative of DNN including the activation function in order to fine-tune the parameters such that the model is optimised. The derivative of the activation function can be simply obtained through chain rule. The derivative of PFTS is written in Equation 4:

$$PFTS'(x) = \begin{cases} \dfrac{1}{1+e^{-x}}(1 - \dfrac{x}{1+e^{-x}}) + \dfrac{x}{1+e^{-x}}, & \text{if } x \geq 0 \\ 0, & \text{if } x < 0 \end{cases} \quad (4)$$

where *x* indicates the input to the derivative of PFTS. The derivative of PFTS is similar to the derivative of FTS; therefore, the derivation steps can be referred from the previous work of Chieng et al. (2018). It is noticeable that the derivative of PFTS has a smooth property at the positive region, instead of treating the positive inputs with a gradient as in ReLU. Meanwhile, at the negative region, the derivative of PFTS maintains the sparsity property as in ReLU to speed up the learning particularly during the backpropagation. The differences between the derivative of PFTS and ReLU can be easily seen in Figure 4.

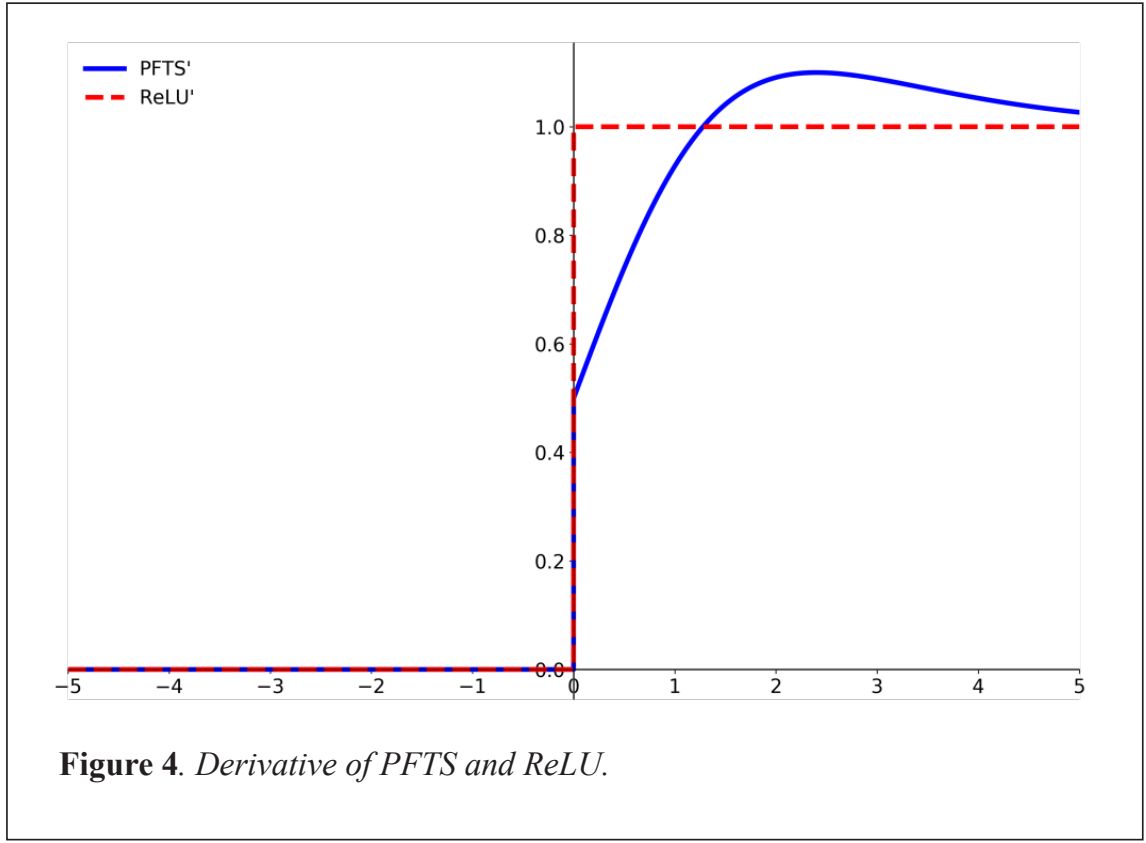

**Figure 4**. *Derivative of PFTS and ReLU.*

## EXPERIMENTAL SETUP

This study aims to investigate the performance of PFTS against ReLU and eight other existing activation functions in terms of classification performance. The details of the activation functions are listed in Table 1, and their corresponding plot is presented in Figure 5. A total of eight standard DNNs with various depths and widths were used to evaluate the activation functions. The configurations of the DNNs are presented in Table 2. All the experiments were implemented using Python with the support of Tensorflow, while GeForce GTX 1060 6GB GPU was utilised to accelerate the computations.

The experiment evaluated the activation functions with different DNNs on the standard Street View House Numbers (SVHN) dataset. SVHN is a 10-class coloured image dataset containing digits ranging from 0 to 9. The dataset has a training set of 73,257 samples and a testing set of 26,032 samples. Each image had been formatted into a fixed 32 x 32 pixels with RGB (red, green, and blue) colour model that had a range of values from 0 to 255. Although SVHN has additional 531,131 samples for the training set, this study did not consider them for the experiment.

Table 1

*Details of The Activation Functions*

| Activation function | Parameter value |
|---|---|
| Equation 1 | None |
| $Swish(x) = \dfrac{x}{1+e^{-x\beta}}$ | $\beta = 1.0$ |
| $Tanh(x) = \dfrac{e^x - e^{-x}}{e^x + e^{-x}}$ | None |
| $LReLU(x) = \begin{cases} x, & if\ x \geq 0 \\ \alpha x, & if\ x < 0 \end{cases}$ | $\alpha = 0.01$ |
| $PReLU(x) = \begin{cases} x, & if\ x \geq 0 \\ \alpha_{train} x, & if\ x < 0 \end{cases}$ | $\alpha_{train} = 0.25$ |
| $Softplus(x) = \ln(1+e^x)$ | None |
| $ELU(x) = \begin{cases} x, & if\ x \geq 0 \\ \alpha(e^x - 1), & if\ x < 0 \end{cases}$ | $\alpha = 1.0$ |
| $FReLU(x) = \begin{cases} x + \beta_{train}, & if\ x \geq 0 \\ \beta_{train}, & if\ x < 0 \end{cases}$ | $\beta_{train} = -0.398$ |
| Equation 2 | $t = -0.20$ |
| Equation 3 | $t_{train} = -0.20$ |

Note: The word "*train*" appearing as subscripted in the activation functions' parameters indicates that they are trainable.

To establish a fair empirical evaluation, all activation functions were tested under identical training conditions. By following the works of Qiu et al. (2018), Chieng et al. (2018), and Ramachandran et al. (2018), the Stochastic Gradient Descent (SGD) (Robbins & Monro, 1951) was adopted as the optimisation technique. The training parameters such as learning rate, dropout rate, batch size, and training epoch were set as 0.01, 0.5, 64, and 50, respectively. Moreover, Xavier initialisation (Glorot & Bengio, 2010) was used for network weight initialisation, while the biases were initialised at 0, and Softmax with cross-entropy loss function was utilised in the output layer.

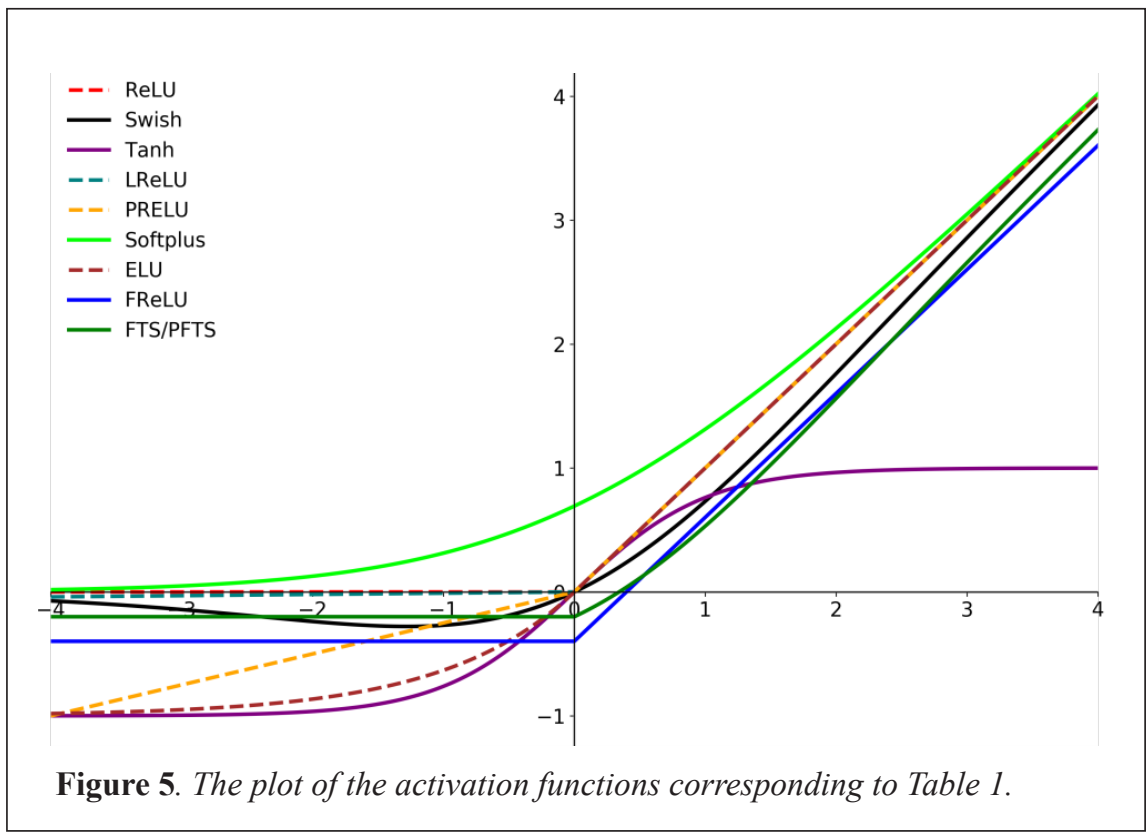

**Figure 5**. *The plot of the activation functions corresponding to Table 1.*

Table 2

*Configuration of The DNNs*

| Network models | Number of hidden layer (including the output layer) | Number of neurons in each layer |
| --- | --- | --- |
| DNN-3A (Pereyra et al., 2017) | 3 | 1024-1024-10 |
| DNN-3B (Zhou & Feng, 2017) | 3 | 1024-512-10 |
| DNN-4 (Scardapane et al., 2017) | 4 | 400-300-100-10 |

(continued)

| Network models | Number of hidden layer (including the output layer) | Number of neurons in each layer |
| --- | --- | --- |
| DNN-5A (Chieng et al., 2018) | 5 | 256-128-64-32-10 |
| DNN-5B (Mohamed et al., 2011) | 5 | 512-512-512-512-10 |
| DNN-5C (Zhou & Feng, 2017) | 5 | 1024-1024-512-256-10 |
| DNN-6 (Chieng et al., 2018) | 6 | 512-256-128-64-32-10 |
| DNN-7 (Chieng et al., 2018) | 7 | 784-512-256-128-64-32-10 |

**Classification Accuracy and Baseline Comparison Score**

Since ReLU is the most widely used activation function in the community, this experiment selected ReLU as a baseline method for performance comparison. By following the same setup in Chieng et al. (2018) and Alcantara (2017), the classification results reported were the mean of five runs. Furthermore, the baseline comparison score (Ramachandran et al., 2018) was aggregated for each activation function, which indicated the number of times each activation function outperformed the ReLU baseline. The classification accuracy results in Table 3 were ranked using the fractional ranking method.

Table 3 shows that PReLU achieved the best performance on all first five DNNs. Nevertheless, the performance dropped drastically in deeper networks, specifically on DNN-6 and DNN-7. This revealed the inconsistency and instability performance of PReLU, particularly in the deeper models. PFTS and Swish, on the other hand, attained the best results on DNN-7 and DNN-6, respectively. Although PFTS did not achieve the best result in the majority of DNNs, it yielded better performance in terms of stability and consistency despite the network configurations. Consequently, PFTS managed to obtain the highest baseline comparison score on eight out of eight DNNs, together with PReLU, FReLU, and FTS. It was worth noting that PFTS outperformed its predecessor, FTS, in all eight network configurations, showing remarkable robustness of the parametric strategy against the non-parametric approach. This finding confirmed the previous works of Qian et al. (2018), Sütfeld et al. (2018), and Jagtap et al. (2019), in which the parametric element in the activation function could generally improve the predictive performance of the network.

Table 3

*Classification Accuracy (%) for SVHN Dataset Using DNNs with Different Activation Functions*

| Activation functions | DNN-3A | DNN-3B | DNN-4 | DNN-5A | DNN-5B | DNN-5C | DNN-6 | DNN-7 | Score |
|---|---|---|---|---|---|---|---|---|---|
| ReLU | 85.75 | 85.41 | 83.29 | 70.19 | 84.14 | 84.90 | 59.32 | 45.97 | - |
| Swish | 85.65 | **85.79** | **84.23** | **81.07** | **84.70** | **85.46** | **82.10** | **81.59*** | 7 |
| Tanh | 81.21 | 80.43 | 73.57 | 57.64 | 75.51 | 77.17 | 55.81 | **54.20** | 1 |
| LReLU | 85.67 | 85.34 | 83.25 | **75.06** | 83.96 | **84.96** | **63.14** | 51.92 | 4 |
| PReLU | **87.00*** | **87.33*** | **86.38*** | **84.45*** | **86.71*** | **87.11*** | **77.68** | 47.01 | 8 |
| Softplus | 83.46 | 83.01 | 79.01 | 63.98 | 79.95 | 80.10 | 47.31 | 19.59 | 0 |
| ELU | 84.87 | 84.77 | 81.59 | **78.78** | 82.47 | 83.07 | **77.02** | **71.74** | 3 |
| FReLU | **86.11** | **86.24** | **85.02** | **82.57** | **85.30** | **85.69** | **82.34** | **78.21** | 8 |
| FTS | **86.00** | **85.96** | **84.46** | **81.29** | **84.86** | **85.46** | **80.26** | **75.64** | 8 |
| PFTS | **86.02** | **86.25** | **85.09** | **82.52** | **85.28** | **85.72** | **83.04*** | **78.99** | 8 |

Note: The best result per configuration is noted with asterisk (*). The values in bold indicate the outperforming results when compared against the ReLU baseline. The "Score" in the last column indicates the "Baseline comparison score".

**Mean Rank**

By following the work of Sütfeld et al. (2018), this experiment additionally evaluated the activation functions from the perspective of mean rank. This was to provide another insight into the performance of the activation functions across the different network configurations. Firstly, the rankings were assigned to the activation functions based on the classification accuracy per network configuration. Then, the mean rank of each activation function across all eight network configurations was computed. The ranking and mean rank are presented in Table 4.

Table 4 shows that the proposed activation function, PFTS, achieved the highest mean rank of 2.25. It was also worth noting that PFTS consistently maintained the top three rankings and outperformed the ReLU baseline across all DNNs. This demonstrated remarkable stability and robustness despite different network configurations. Although PReLU consistently maintained the first place in the first six DNNs, the ranking dropped drastically in the last two deeper networks, achieving the mean rank of 2.38. On the other hand, the ReLU baseline only obtained a mean rank of 6.88, which was ahead of ELU, Softplus, and Tanh. In general, these once again siginified that the parametric activation functions could significantly enhance network performance over the non-parametric ones.

Table 4

*The Mean Rank of the Activation Functions Across All Eight Network Configurations*

| Activation functions | DNN-3A | DNN-3B | DNN-4 | DNN-5A | DNN-5B | DNN-5C | DNN-6 | DNN-7 | Mean rank |
| --- | --- | --- | --- | --- | --- | --- | --- | --- | --- |
| ReLU | 5 | 6 | 6 | 8 | 6 | 7 | 8 | 9 | 6.88 |
| Swish | 7 | 5 | 5 | 5 | 5 | 4.5 | 3 | 1 | 4.44 |
| Tanh | 10 | 10 | 10 | 10 | 10 | 10 | 9 | 6 | 9.38 |
| LReLU | 6 | 7 | 7 | 7 | 7 | 6 | 7 | 7 | 6.75 |
| PReLU | 1 | 1 | 1 | 1 | 1 | 1 | 5 | 8 | 2.38 |
| Softplus | 9 | 9 | 9 | 9 | 9 | 9 | 10 | 10 | 9.25 |
| ELU | 8 | 8 | 8 | 6 | 8 | 8 | 6 | 5 | 7.13 |
| FReLU | 2 | 3 | 3 | 2 | 2 | 3 | 2 | 3 | 2.50 |
| FTS | 4 | 4 | 4 | 4 | 4 | 4.5 | 4 | 4 | 4.06 |
| PFTS | 3 | 2 | 2 | 3 | 3 | 2 | 1 | 2 | **2.25** |

Note: The lower the mean rank value, the higher the rank. The highest mean rank is in bold.

**Discussion**

The outcomes from the experimental results showed that PFTS consistently achieved stable performance and outperformed the ReLU baseline on average. PFTS had several notable differences that fulfilled the criteria of good activation function as presented in the Criteria of Good Activation Function section. Firstly, it allowed negative activation particularly during the forward propagation to provide additional "clues" or "hints" for a network to make sense or reasoning about the instances. Meanwhile, its derivative preserved the sparseness as in ReLU to accelerate learning. Secondly, its negative region mitigated the bias shift effect in the network layers. Assuming the input $x \sim N(0,1)$, the mean activation of PFTS and ReLU were 0.074 and 0.357, respectively. Apparently, PFTS had a mean activation that was much closer to zero as compared to ReLU. This further improved the learning efficiency of the networks. Thirdly, PFTS increased the learning capability and facilitated the flexibility of DNNs via parametric strategy, and therefore, a different activation response could be learned for different layers. In addition to this, a greater non-linear representation capability was endowed to the network when a parametric activation function was used. Lastly, PFTS possessed a curvature property in the activation function where ReLU did not own. This non-linear property allowed PFTS to manifest higher non-linear approximation power during the training and thereby improving the predictive performance of the networks.

## CONCLUSION

In this paper, a Parametric Flatten-T Swish (PFTS) activation function is presented. This activation function uses parametric strategy to learn its activation response from the network layers based on the inputs. Unlike most of the parametric approaches where the slope segments are learned, PFTS learns the hinge point of the function without manipulating its original shape. It introduces better flexibility, stability, expressivity, robustness as well as enhances network predictive performance. The experiments compared PFTS with other existing activation functions, where ReLU was served as the baseline method. The results revealed that PFTS consistently achieved a stable performance and outperformed the ReLU baseline across eight different DNNs on image classification tasks. Specifically, PFTS yielded improvement of 0.31%, 0.98%, 2.16%, 17.72%, 1.35%, 0.97%, 39.99%, and 71.83% on DNN-3A, DNN-3B, DNN-4, DNN-5A, DNN-5B, DNN-5C, DNN-6, and DNN-7, respectively, as compared to the ReLU baseline on the SVHN image dataset. In terms of mean rank, PFTS achieved the highest mean rank of 2.55, outperforming all the other existing activation functions and making it a feasible alternative to ReLU and other existing activation functions. In future work, PFTS is suggested to be enhanced by allowing the bi-directional adaptivity in a greater range of flexibility and expressivity. Such enhancement will be tested on several more challenging datasets using different DNN configurations.

## ACKNOWLEDGEMENT

The work conducted in this study was supported in part of the Geran Penyelidikan Pascasiswazah (GPPS) Vot U817 under the Office for Research, Innovation, Commercialisation and Consultancy Management (ORICC) of Universiti Tun Hussein Onn Malaysia (UTHM).